\documentclass{article}

\usepackage{arxiv}

\usepackage[utf8]{inputenc} 
\usepackage[T1]{fontenc}    
\usepackage{hyperref}       
\usepackage{url}            
\usepackage{booktabs}       
\usepackage{amsfonts}       
\usepackage{nicefrac}       
\usepackage{microtype}      
\usepackage{lipsum}		
\usepackage{graphicx}
\usepackage[numbers]{natbib}
\usepackage{doi}
\usepackage{authblk}

\usepackage{amsfonts}
\usepackage{amsmath}

\usepackage{bm}
\usepackage{bbm}
\usepackage{caption}
\usepackage{subcaption}
\usepackage{booktabs}
\usepackage{xcolor}
\usepackage{url}

\newtheorem{theorem}{Theorem}[section]

\newtheorem{remark}{Remark}

\newtheorem{example*}{Example}

\newcommand{\parentheses}[1]{\left( #1 \right)}

\newcommand{\brackets}[1]{\left[ #1 \right]}

\newcommand{\dprev}{\Delta_{ij}(\eta_k)}
\newcommand{\dnext}{\Delta_{ij}(\eta_{k+1})}
\newcommand{\norm}[1]{||#1||}
\newcommand{\dotproduct}[2]{\langle#1,#2\rangle}

\newcommand{\iidsim}{\overset{iid}{\sim}}

\newcommand{\Ecal}{\mathcal{E}}
\newcommand{\Ucal}{\mathcal{U}}

\newcommand{\Rpos}{\Rbb_{+}^*}

\newcommand{\deltaequal}{\overset{\Delta}{=}}

\newcommand{\pz}{\Pbb_{Z}}
\newcommand{\pp}{\Pcal \Pcal}

\def\eqref#1{equation~\ref{#1}}

\def\1{\bm{1}}

\def\rvz{{\mathbf{z}}}

\def\rmY{{\mathbf{Y}}}

\def\vz{{\bm{z}}}

\def\mI{{\bm{I}}}

\def\mN{{\bm{N}}}

\def\mY{{\bm{Y}}}

\DeclareMathAlphabet{\mathsfit}{\encodingdefault}{\sfdefault}{m}{sl}
\SetMathAlphabet{\mathsfit}{bold}{\encodingdefault}{\sfdefault}{bx}{n}

\def\Ccal{{\mathcal{C}}}

\def\Ecal{{\mathcal{E}}}

\def\Lcal{{\mathcal{L}}}
\def\Mcal{{\mathcal{M}}}
\def\Ncal{{\mathcal{N}}}

\def\Pcal{{\mathcal{P}}}

\def\Tcal{{\mathcal{T}}}
\def\Ucal{{\mathcal{U}}}

\def\Zcal{{\mathcal{Z}}}

\def\Nbb{{\mathbb{N}}}

\def\Rbb{{\mathbb{R}}}

\newcommand{\E}{\mathbb{E}}

\newcommand{\ptitle}[1]{\noindent{\bf #1}.}

\title{Gaussian Embedding of Temporal Networks}

\author[ \space 1]{Raphaël Romero \thanks{Corresponding author (e-mail:
	raphael.romero@ugent.be)}} 
\author[1]{Jefrey Lijffijt}
\author[2]{Riccardo Rastelli}
\author[3,4]{Marco Corneli}
\author[1]{Tijl De Bie}

\affil[1]{Ghent University}
\affil[2]{University College Dublin, School of Mathematics and Statistics, Dublin, Ireland}
\affil[3]{Université Côte d’Azur, CNRS, Laboratoire CEPAM, Nice, France}
\affil[4]{Université Côte d’Azur, CNRS, INRIA, Laboratoire LJAD, Nice, France}

\date{}

\renewcommand{\headeright}{}
\renewcommand{\undertitle}{}
\renewcommand{\shorttitle}{\textit{arXiv} Template}

\begin{document}
\maketitle

\begin{abstract}
	Representing the nodes of continuous-time temporal graphs in a low-dimensional latent space has wide-ranging applications, from prediction to visualization. Yet, analyzing continuous-time relational data with timestamped interactions introduces unique challenges due to its sparsity. Merely embedding nodes as trajectories in the latent space overlooks this sparsity, emphasizing the need to quantify uncertainty around the latent positions. In this paper, we propose TGNE (\textbf{T}emporal \textbf{G}aussian \textbf{N}etwork \textbf{E}mbedding), an innovative method that bridges two distinct strands of literature: the statistical analysis of networks via Latent Space Models (LSM)\cite{Hoff2002} and temporal graph machine learning. TGNE embeds nodes as piece-wise linear trajectories of Gaussian distributions in the latent space, capturing both structural information and uncertainty around the trajectories. We evaluate TGNE's effectiveness in reconstructing the original graph and modelling uncertainty. The results demonstrate that TGNE generates competitive time-varying embedding locations compared to common baselines for reconstructing unobserved edge interactions based on observed edges. Furthermore, the uncertainty estimates align with the time-varying degree distribution in the network, providing valuable insights into the temporal dynamics of the graph. To facilitate reproducibility, we provide an open-source implementation of TGNE at \url{https://github.com/aida-ugent/tgne}.
\end{abstract}

\keywords{Temporal Graphs \and Latent Space Models \and Variational Inference \and Representation Learning \and Dimensionality Reduction \and Networks}

\section{Introduction}
Continuous-time temporal networks arise from various sources. They have successfully been used to study communication patterns, epidemic spreading, and neuron firing, to name a few.
Moreover, the interaction data is often available with a high level of detail, making it possible to model the time dimension as continuous.
In that setting, any temporal interaction network can be viewed as the realization of a collection of edge-specific point processes, enabling the use of statistical methods to study the dynamics of the interactions \cite{NaokiMasuda}.

Latent Space Models for graphs \cite{Hoff2002} are an important class of probabilistic models, where each node in the graph is embedded into a latent space, and the probabilities of links between nodes are independently distributed according to a notion of distance between node embeddings. Such dyad-independent models allow one to reliably infer unobserved node-level information based on the observation of links between the nodes. The learned node embeddings can then be used directly for downstream tasks such as clustering or link prediction \cite{Kang2018}.
Latent Space Models have been extended to make them applicable to a variety of network types, and publicly available packages allow analyzing a broad range of relational data \cite{krivitskyFittingPositionLatent2008a}.

For continuous time temporal graphs, however, where each interaction is allowed to occur at any time stamp, the translation of Latent Space Models has not been fully explored yet. Indeed, for this type of data, the point process nature of the edge-level variables that generate the data does not allow parameterizing the nodes into simple embedding vectors, but instead in theory would require a full trajectory of embeddings.
To cope with these limitations, recently the Continuous Latent Position Model (CLPM) \cite{rastelliContinuousLatentPosition2021} was introduced, which sticks with the Point Process nature of the data, while assuming the latent trajectories of embeddings to be piece-wise linear in order to derive a fully parametric model.
In \cite{rastelliContinuousLatentPosition2021}, the authors propose Maximum A Posteriori inference to estimate the latent trajectories based on a history of dyadic interactions. However, the authors do not consider estimating the uncertainty of the estimated trajectories. Given the sparsity of typical interaction networks, such uncertainty may be large and vary across nodes, dyads and time. Yet, understanding this uncertainty may be crucial in many applications.
To meet this need, in this work we present TGNE (Temporal Gaussian Network Embedding), which appeals to Bayesian inference to capture a time-varying notion of uncertainty of the model on the latent position. 
While still allowing one to visualize temporal networks in a low-dimensional space, TGNE additionally allows one to gauge the uncertainty around the latent positions in a natural and rigorous manner.

Our contributions can be summarized as follows.
\begin{itemize}
    \item We propose TGNE: a variational approach to inference in the CLPM model that allows calculating trajectories of Gaussian in a latent space, given a history of interactions. 
    \item We carry out an exploratory analysis of simulated and real-world datasets using the obtained dynamic embeddings.
    \item We evaluate the uncertainty learned through the variational approximation of the posterior.
    \item We assess to what extent this method can be used to reconstruct missing events in the temporal network.
\end{itemize}

The paper is organized as follows. In Section \ref{related_work} we discuss related work. In Section \ref{prelim} we provide a Point Process perspective to Continuous-Time Temporal Networks and introduce the CLPM model in light of these definitions. In Section \ref{method}, we detail TGNE, and discuss its scalability. In Section \ref{experiments} we detail and discuss the results of our experiments. Finally, in Section \ref{discussion} we outline potential extensions of the proposed work.

\section{Related Work}
\label{related_work}
Our work builds on previous work on Latent Space and Point Process Modelling of (Temporal) Graphs. 

\ptitle{Latent Space Models}
Since Hoff's seminal paper \cite{Hoff2002}, Latent Position Models (LPMs) have been extensively studied and expanded to various types of graphs, including weighted graphs and dynamic graphs \cite{sarkarDynamicSocialNetwork2005, Kim2018}. The Continuous Latent Position Model (CLPM) \cite{rastelliContinuousLatentPosition2021} further extends this line of research to continuous-time temporal graphs, where the latent positions of nodes are assumed to follow piece-wise linear trajectories in a latent space. Our work builds upon this model and describes a Bayesian approach for estimating the latent trajectories.

\ptitle{Dynamic Graph Layout and Diachronic Embedding}
Dynamic Graph Layout \cite{xuRegularizedGraphLayout2013} aims to find embedding configurations that not only represent the structural information of the graph but also maintain coherence over time. Similarly, Goel et al. \cite{goelDiachronicEmbeddingTemporal2019}  propose Diachronic Embedding, which enables embedding nodes from a knowledge graph into a coherent sequence of latent embeddings for temporal knowledge graph completion. As detailed in the method section, we also enforce temporal coherence by specifying a Gaussian Random Walk prior distribution over the latent trajectories.

\ptitle{Gaussian Graph Embedding}
Recent work has explored the idea of embedding nodes in a graph as Gaussian-distributed points in a latent space, with extensions to dynamic graphs \cite{bojchevskiDeepGaussianEmbedding2018,xuDynG2GEfficientStochastic2022}. However, the main focus of this line of research has primarily been on forecasting in discrete-time temporal graphs. In contrast with this, our work aims to provide a Bayesian dimensionality reduction method specifically tailored for temporal graphs in \emph{continuous-time}.

\ptitle{Point Process Models for graphs}
Point Process Modeling of Temporal Graphs, particularly using Hawkes Process models, has emerged as a vibrant field of research
\cite{Arastuie2019,huangMutuallyExcitingLatent2022,passinoMutuallyExcitingPoint2022a,yangDecouplingHomophilyReciprocity}. These models characterize the changing rates of events in a network based on hidden representations.
However, in existing models, the interaction rates are typically modulated by static representations of the nodes. Few efforts have been dedicated to combining these Point Process decoders with continuous-time representations, which is a key aspect of our work.

\ptitle{Temporal Graph Neural Networks}
 Automatically learning time-varying node feature vectors from time-stamped relational data through encoder-decoder architectures is a very active field of research, as surveyed in \cite{kazemiRepresentationLearningDynamic}. 
 Such architectures are evaluated on two classes of tasks. \emph{Interpolation} aims at reconstructing past events, and is mostly evaluated on knowledge graphs at a typically low time-resolution. On the other hand, Temporal GNNs are typically evaluated on their ability to \emph{extrapolate} to the future. In contrast, the method proposed in this paper is a dimensionality-reduction method aimed at capturing both the structure of the graph at a user-specified resolution, along with uncertainty on the latent node representations.

 \section{Preliminaries}
 \label{prelim}
 In this section, Temporal Networks are defined from a Point Process point of view. Then the Poisson Process is defined: a particular type of point process that is used in this paper as a generative distribution of the data. Finally, we summarize the Continuous Latent Position Model (CLPM) in light of this theoretical background.
 \subsection{Notations}
 
 \ptitle{Continuous-time Temporal Networks}
 Let $\Ucal$ denote a set of nodes, and $\Ecal \subset \Ucal \times \Ucal$ a set of possible edges.
 In the current work, a \textbf{Temporal Network} is defined as a time-ordered sequence of relational events
 $\Tcal([0,1]) = \{w_m=(i_m,j_m, t_m)|m=1,\ldots,M\},$ where $ M$ is the number of events, $0<t_1<\cdots<t_M<1$ is an ordered sequence of pairwise distinct, positive time stamps, 
 and $i_m$ and $j_m$ are the source and destination nodes respectively. The time-stamps are normalized to the interval $[0,1]$. For any node pair $i,j \in \Ecal$, and 
 for any $0\leq a<b \leq 1$, we denote $\Tcal_{ij}([a,b])$ the set of interaction times between $i$ and $j$ that occur in the interval $[a,b]$.
 For each $(i,j)\in \Ecal$ we define the function $t\mapsto \mY_{ij}(t)\in \Nbb$ that counts the number of interactions between $i$ and $j$ before time $t$.
 We assume that the edge-level counting functions are samples from simple point processes, and we will denote $t\mapsto \rmY_{ij}(t)$ the \emph{counting process} generating the time function $t\mapsto \mY_{ij}(t)$.

 \ptitle{Poisson Processes}
 A Poisson Point Process on the interval $[0,1]$ is a random variable that, when sampled from, yields a set of arrival times $t_1,\ldots,t_m$. 
 Such a random variable is governed by its rate function $\lambda:[0,1]\mapsto\Rpos$, defined such that for any interval $[a,b]\subset[0,1]$, the expected number of arrival times that fall into $[a,b]$ is given by the rate measure:
 \begin{align*}
     \Lambda([a,b])\deltaequal\E[\rmY(b)-\rmY(a)]=\int_{a}^b \lambda(s)ds.
 \end{align*}
 In other words, $\lambda(t)$ can be viewed as the expected number of events occurring in the interval $[t, t+dt[$. 
 For a given rate function $\lambda:[0,1]\rightarrow\Rbb_{+}^*$, we will write $\rmY\sim \pp(\lambda)$ to express that $\rmY$ follows a Poisson Process distribution with rate function $\lambda$. 
 The likelihood of observing the arrival times $t_1,...,t_m$ under a Poisson Process of rate function $\lambda$ is:
 $p(\{t_1,...,t_m\};\lambda)=\exp(-\Lambda([0,1]))\prod\limits_{i=1}^m \lambda(t_i).$
 \begin{remark}
     This can also be written in the following exponential family form:
     $$p(\{t_1,...,t_m\};\lambda)= \exp\left(\int_0^1\log\lambda(s)d\mY(s) - \Lambda([0,1])\right),$$
 where $\mY(t)=\sum_{i=1}^m\mathbbm{1}_{t_i<t}$ is the counting function representing the arrival times and 
 $\int_0^1\log\lambda(s)d\mY(s)$ is the Stieltjes integral of the log rate with respect to $\mY$. 
 While the natural parameter of this exponential family is the function $s\mapsto \log\lambda(s)$, $\mY$ can be interpreted as the sufficient statistics. Thus the canonical link function is the $\log$ in that case. The second term in the exponential is in turn the log-partition function of the distribution. 
 This exponential form makes the Poisson Process a natural candidate as a generative model in a continuous-time extension of the Latent Space Distance Model \cite{Hoff2002}.
 \end{remark}

\subsection{The Continuous-time Latent Space Model}

\begin{figure*}[t!]
    \centering
    \includegraphics[width=\linewidth]{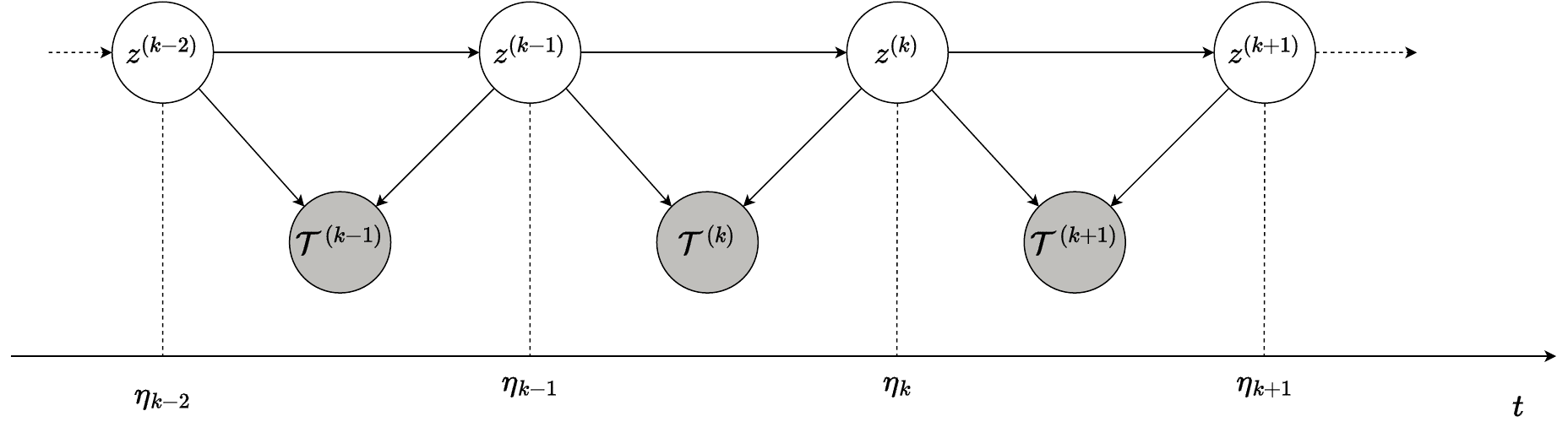}
    \caption{Probabilistic Graphical Model summarizing the CLPM.
    $\Tcal^{(k)} = \{(i_m, j_m, t_m)\in\Tcal|t_m \in I_k\}$ is the history of interactions happening in the time interval $I_k=[\eta_{k-1}, \eta_k]$.
    $z^{(k)}$ are the snapshots of latent positions at time $\eta_k$.
    The chunks of history $\Tcal^{(k)}$ are conditionally independent given the latent positions at the boundaries of the interval $I_k$, namely $z^{(k-1)}$ and $z^{(k)}$}
    \label{fig:clpm_pgm}
\end{figure*}

\ptitle{General Summary}
\label{sec:clpm}
The Continuous-time Latent Space Model (CLPM) can be summarized as follows. 
Let $\Mcal$ be an embedding space (typically $\Mcal=\Rbb^d$ with $d$ a small latent space dimension), $\Zcal=\Ccal([0,1], \Mcal)$ the set of continuous trajectories in that latent space and $\Ccal([0,1], \Rpos)$ the set of positive continuous functions on $[0,1]$.
\renewcommand{\pz}{p_{\vz}} 
Let $\pz$ be a prior distribution over $\Zcal$.
The model supposes that the edge-level interaction times are generated independently conditioned on the latent trajectories, using:
\begin{align*}
    \forall i, \rvz_i &\sim p_{\vz},\\
    \forall i,j, \rmY_{ij}|\vz_i,\vz_i &\sim \pp(f(\vz_i,\vz_j)),
\end{align*}
%
%
%
%
where $f:\Zcal \times \Zcal \rightarrow \Ccal([0,1], \Rpos)$ is a \emph{similarity function} that maps any two trajectories $\vz,\vz'$ to a positive rate function $f(\vz,\vz')=(t\mapsto f_{\vz,\vz'}(t))$.
Examples of such a model include the distance model, where the similarity function is given by $f_{\vz,\vz'}(t)=\exp(\beta-||z_i(t)-z_j(t)||^2)$ and the dot product model corresponding to $f_{\vz,\vz'}(t)=\exp(\beta+\left<z_i(t),z_j(t)\right>)$. 

\ptitle{A Piecewise Linear Assumption}
Rastelli and Corneli \cite{rastelliContinuousLatentPosition2021} propose to constrain the trajectories to be piece-wise linear to make the model tractable. 
The observation interval $[0,1]$ is partitioned into a set of $K$ intervals $I_1,...,I_K$ with 
$I_k=[\eta_{k-1}, \eta_k]$, resulting in $K+1$ cut-points $0=\eta_0<\eta_1<\eta_2\ldots<\eta_{K}=1$. 
The latent trajectories are then assumed to be linear on each interval $I_k$. Formally, for each node $i$, interval $k$ and coefficient $s\in[0,1]$, the latent position $\vz_i$ at time $t=(1-s)\eta_{k-1}+s\eta_{k} \in I_k$ is:
 \begin{align*}
      \vz_i((1-s)\eta_{k-1} + s\eta_{k}) &= (1-s)\vz_i(\eta_{k-1}) + s\vz_i(\eta_{k}).    
 \end{align*}
 Thus, the i-th trajectory is fully determined by its successive positions at the change points $\{\vz_i(\eta_k)|k=0,\ldots,K\}$, which means that only $(K+1)\times d$ variables are needed to describe it. 
 The positions at the cut-points are referred to as \textbf{critical points} in the following, and denoted $\vz_i^{(k)} \deltaequal \vz_i(\eta_k)$.
 
 

\ptitle{Log-Likelihood of the CLPM}
For each node pair $(i,j)$ and each interval $I_k$, let $\mY_{ij}^{(k)}$ be the number of interactions between $i$ and $j$ that occur in the interval $I_k$.
The associated random variables $\rmY_{ij}^{(k)}$ are independent across node-pair and intervals, conditioned on the latent trajectories. Moreover, $\mY_{ij}^{(k)}$ only depends on the latent positions of $i$ and $j$ at the boundaries of the interval $I_k$, namely $\{\vz_i^{(k-1)},\vz_i^{(k)}, \vz_j^{(k-1)},\vz_j^{(k)}\}$. The independence structure of the CLPM is summarized in Figure \ref{fig:clpm_pgm}.
The negative log-likelihood conditioned on the latent positions thus decomposes as follows:
\begin{align*}
    \log p(\mY|\vz) 
    = -\sum_{i\neq j} \sum_{k=1}^K \log p(\mY_{ij}^{(k)}|\vz_i^{(k-1)},\vz_i^{(k)}, \vz_j^{(k-1)},\vz_j^{(k)}).
\end{align*}
The terms in this decomposition are the following Poisson Process log-likelihoods:
\begin{align*}
    &-\log p(\mY_{ij}^{(k)}|\vz_i^{(k-1)},\vz_i^{(k)}, \vz_j^{(k-1)},\vz_j^{(k)})\\
    &=\Lambda_{ij}(I_k) -\sum_{t_{ij} \in \Tcal_{ij}(I_k)} \log(\lambda_{ij}(t_{ij})).    
\end{align*}

While the second term can be calculated directly from the parameters, the cumulative rate $\Lambda_{ij}(I_k)$ is more difficult to evaluate. We describe two options for calculating this term: the closed form already described in \cite{rastelliContinuousLatentPosition2021}, and an approximate form based on a Riemann sum.

\ptitle{Closed form \cite{rastelliContinuousLatentPosition2021}}
In the particular case of the Euclidean Distance model, the cumulative rate adopts the following closed form:

\begin{theorem}
We have $$\Lambda_{ij}(I_k) = (\eta_{k-1}-\eta_{k})\exp(\beta - a) \sqrt{2\pi} 
\Big[
    \Phi(\frac{1-\mu}{\sigma}) - \Phi(-\frac{\mu}{\sigma})
\Big]$$
where:
\begin{itemize}
    \item $\Phi:t \mapsto \frac{1}{\sqrt{2\pi}}\int_{-\infty}^u e^{-\frac{u^2}{2}}du$ is the standard Normal $\Ncal(0,1)$ cumulative distribution function.
    \item $\sigma=\frac{1}{\sqrt{2}||\dprev-\dnext||}$.
    \item $\mu=\frac{\langle\dprev, \dprev-\dnext\rangle}{||\dprev-\dnext||}$.
    \item $a=||\dprev||^2-
    \frac{\dotproduct{\dprev}{\dprev-\dnext}^2}{
        \norm{\dprev-\dnext}^2
    }$
    \item $\Delta_{ij}(\eta_k) = z_i(\eta_k) - z_j(\eta_k)$.
\end{itemize}
\end{theorem}

While a proof of this theorem is provided in \cite{rastelliContinuousLatentPosition2021}, in the supplementary material we provide an alternative proof that can be reproduced for any case where the log of the rate can be expressed as a second-order spline in time, i.e. such that its expression on each subsequent interval is a second-order polynomial.

\ptitle{Riemann approximation of the cumulative rate} 
In the case where the rate function is not a second-order spline, we propose to approximate it simply using a Riemann sum: 
\begin{align*}
    \Lambda_{ij}(I_k) = \int_{\eta_{k-1}}^{\eta_k} \lambda_{ij}(s)ds \approx \frac{1}{R}\sum_{r=1}^{R} \lambda_{ij}(\eta_{k-1} + \frac{r-1}{R}  \eta_k)
\end{align*}
where $R$ is a pre-specified resolution parameter.
This approximation allows implementing an inference procedure agnostic to the type of similarity function used. For instance, using this approximation makes it easy to consider different latent geometries such as hyperbolic or spherical embeddings.

\section{Method}
\label{method}
In this section, we provide an overview of the proposed TGNE approach for performing Bayesian inference on the latent critical points, given a history of interactions.

\subsection{Prior distribution}
\label{prior}
To reflect time continuity in the latent trajectories of the CLPM, a prior distribution is needed. The Gaussian Random Walk prior introduced in \cite{rastelliContinuousLatentPosition2021} biases the inference towards time-coherent and reasonably scaled configurations, promoting slowly evolving trajectories while faithfully representing the network's structure.
This prior is defined for any node $i\in \{1,\ldots,n\}$ and time step $k\in \{0 ,\ldots, K\}$, as the cumulative sum of independent Gaussian increments:
\begin{align*}
    z_i(\eta_k) = \tau_0 \epsilon_0 + \sum_{l=1}^k  \sqrt{\eta_{l}-\eta_{l-1}}\tau\epsilon_l,
\end{align*}
where $\epsilon_i\sim \Ncal(0,I_d)$.
The initial scale $\tau_0$ controls the overall spread of the latent trajectories in the embedding space.
The transition scale parameter $\tau$ governs the amount of allowed variation between consecutive time steps.
Additionally, the variance of the Gaussian increments increases linearly with the step size $\eta_{k+1}-\eta_k$.
Note that by taking infinitely small step sizes, this prior converges to a Brownian Motion in the embedding space. 
In our implementation, we choose a constant step size $\eta_{k+1}-\eta_k = \frac{1}{K}$, where $K$ is the number of steps. 
Moreover, we select an initial scale equal to the transition scale: $\tau_0=\tau$.
This yields two hyperparameters: the scale $\tau$ and the number of change points (ticks) $K$.
\subsection{Variational Inference on the Critical Points}
\newcommand{\zparam}{(z_i(\eta_k))_{i\in[n], k\in[K]}}
\newcommand{\vardist}{q_{\phi}}
\newcommand{\posterior}{p(.|\mY)}
\newcommand{\zik}[1][k]{\vz_i^{(#1)}}
\newcommand{\muik}[1][k]{\mu_i^{(#1)}}
\newcommand{\sik}[1][k]{\sigma_i^{(#1)}}
The objective of TGNE is to evaluate the intractable posterior distribution $p(\vz|\mY)\propto p(\mY|\vz)p(\vz)$ given the data $\mY$. To achieve this, we use a mean-field variational approach, where we define the following variational distribution that factorizes over nodes and change points as a product of independent Normal distributions:

\begin{align}
\label{mean-field}
q_{\phi}(\vz) =\prod_{i=1}^n\prod_{k=0}^K \Ncal(\zik; \muik, (\sik)^2\mI_d),
\end{align}

We aim to minimize the Kullback-Leibler divergence $KL(q_{\phi}||p(.|\mY))$ between the variational distribution and the posterior. This is equivalent to minimizing the negative Evidence Lower Bound (ELBO):

\begin{align*}
\Lcal(\phi) = KL(q_\phi||p(.|\mY)) - \E_{\vz\sim q_{\phi}}[\log(p(\mY|\vz))],
\end{align*}
The KL Divergence term can be written as shown theorem \Ref{theorem:kl}, and proved in Appendix \Ref{klproof}.

\begin{theorem}
    \label{theorem:kl}                      
\begin{align*}
    KL(q_{\phi}||p(.|\mY)) 
    &= 
    \sum_{i=1}^{n} 
    \left[
        \frac{||\muik[0]||^2}{2\tau_0^2} \right.\\
        &+d\sum_{k=0}^K
        \left(
            \log(\frac{\sik}{\tau}) 
            + 
            \frac{\tau^2}{(\sik)^2}-\frac{1}{2}
            \right)
            \\
        &
        \left.
        +\sum_{k=1}^K
        \frac{
            ||\muik-\muik[k-1]||^2+ (\sik[k-1])^2
        }{
            2\tau^2
            }
    \right]. 
\end{align*}
\end{theorem}


Following common practices in variational inference, the expected log-likelihood term is approximated using a single Monte-Carlo sample $\tilde{\vz}\sim q_{\phi}$:

\begin{align*} 
    \E_{\vz\sim q_{\phi}}[\log(p(\mY|\vz))] \approx \log(p(\mY|\tilde{\vz})).
\end{align*} 

Reparameterization \cite{Kingma_Welling_2014} allows backpropagating through the latter sampling operation, by mapping standard Normal-distributed samples to the latent space through an invertible function of the variational parameters. It is used here to obtain the following differentiable loss, which can be optimized using standard gradient descent algorithms such as ADAM \cite{Kingma2015}:

\begin{equation}
    \label{full_loss}
    \begin{aligned}
    &\Lcal(\phi)\approx
    \overbrace{
        \sum_{i,j}\sum_{k=1}^{K}\Lambda_{ij}(I_k) -
        \sum_{t_{ij} \in \Tcal_{ij}(I_k)} \log(\lambda_{ij}(t_{ij}))
    }^{\text{Calculated using a single sample $\tilde{\vz}\sim q_{\phi}$}}
    \\
    &+
    \sum_{i=1}^{n} 
    \left[
        \frac{||\muik[0]||^2}{2\tau_0^2} \right.
        +d\sum_{k=0}^K
        \left(
            \log\parentheses{\frac{\sik}{\tau}} 
            + 
            \frac{\tau^2}{(\sik)^2}-\frac{1}{2}
            \right)
            \\
        &
        \left.
        +\sum_{k=1}^K
        \frac{
            ||\muik-\muik[k-1]||^2+ (\sik[k-1])^2
        }{
            2\tau^2
            }
    \right].
    \end{aligned}
\end{equation}

\subsection{Effect of the hyperparameters}
The proposed method has four hyperparameters: the dimension $d$, the number of change points $K$, the initial scale $\tau_0$, and the scale $\tau$.
The number of change points $K$ controls the time resolution of the latent trajectories. It should be adapted to how frequently we expect the nodes' states to change in our dataset.
The initial scale $\tau_0$ controls the scale of the initial latent positions $\zik[0]$. 
Finally, the scale $\tau$ is a temperature parameter that controls the deviation of the latent positions between frames, namely $\norm{\zik[k+1] - \zik}$. To illustrate its effect, in Figure \ref{fig:sbm} it can be seen that for $\tau=50.0$ the frames are not constrained to be close to each other, and the latent positions can change drastically between frames. On the other hand, for $\tau=1.0$, the latent positions are constrained to be close to each other, and the frames are more similar to each other.

\subsection{Implementation}
We implemented our method in Pyro, a Pytorch-based probabilistic programming language \cite{binghamPyroDeepUniversal}. This effect handler-oriented programming language allows one to define the model as a Python function. The execution trace of the function can then be read and decorated by effect handlers, allowing one to define high-level probabilistic operations such as conditioning, or performing Stochastic Variational Inference. 
To optimize the variational parameters $\phi$ and the bias term $\beta$, we use the ADAM algorithm \cite{Kingma2015} with learning rates $\gamma=0.01$ and $\gamma=0.00001$ respectively.

\subsection{Scalability}
We discuss two strategies to scale the method to networks with a large number of nodes: node-batching and negative sampling.

As the log likelihood term is a sum of terms over all source nodes, \textbf{node-batching} can be implemented by computing the loss and gradients on a subset of the nodes at each iteration, and then averaging the gradients over the whole dataset.

The log-likelihood decomposes as a sum of contributions from \textbf{positive} node pairs (interacting at least once) and \textbf{negative} node pairs that never interact. 
However, most of the node pairs in the network never interact, and thus the information conveyed by the negative pairs is redundant. This opens up the possibility of \textbf{negative sampling} which may dramatically speed up inference on networks with many nodes. We propose the following strategy, akin to the case-control approximate likelihood introduced in \cite{rafteryFastInferenceLatent2012c}: for each node $i$, we sample $K$ nodes $j$ such that $(i,j)$ never interact in the network. We denote $\Pcal(i)$ the set of nodes that connect with $i$ at least once in the event history, and $\Ncal(i)$ a random subset of the set of nodes that never connected with $i$. The log-likelihood can be approximated as:

\begin{align*}
    \log(p(\mY|\tilde{\vz}))
    \approx &\sum_{i\in \Ucal} 
    \brackets{
        \sum_{j\in \Pcal(i)} \int_{0}^{1} \lambda_{ij}(s)ds - \sum_{\tau\in \Tcal_{ij}} \log(\lambda_{ij}(\tau))
    }\\
    &+ 
    \frac{|\Ucal \setminus  \Pcal(i)|}{|\Ncal(i)|}
    \brackets{
        \sum_{j\in\Ncal(i)}\int_{0}^{1} \lambda_{ij}(s)ds
    }
\end{align*}

The efficacy of the negative sampling approach can be further enhanced by tailoring the selection of negative samples to each specific interval. This refinement is intended to increase the complexity of the negative samples, leading to a more accurate estimation of the gradient. Nevertheless, in our proposed work, we employ a single set of negative samples for all intervals, as this approach already yields satisfactory results on the considered datasets.

\section{Experiments}

\begin{figure*}[h]
    \begin{subfigure}[!]{\textwidth}
        \centering
        \includegraphics[width=\linewidth]{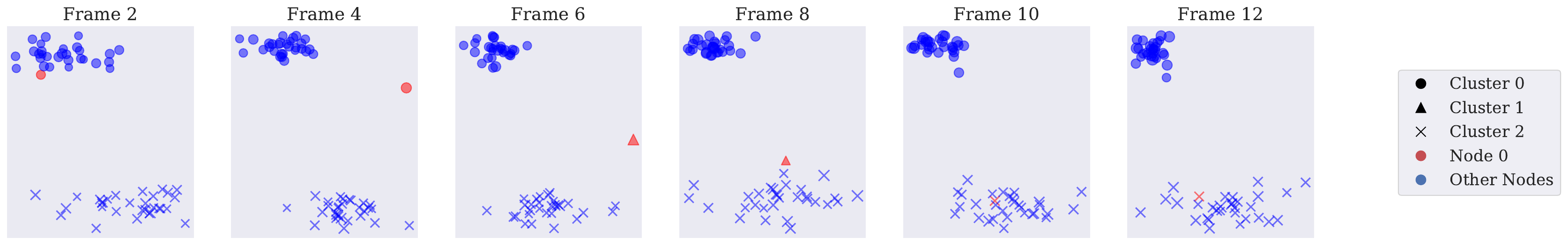}
        \caption{$K=15,\tau=1.0$} 
        \label{fig:sbm:a}
    \end{subfigure}
    \begin{subfigure}[!]{\textwidth}
        \centering
        \includegraphics[width=\linewidth]{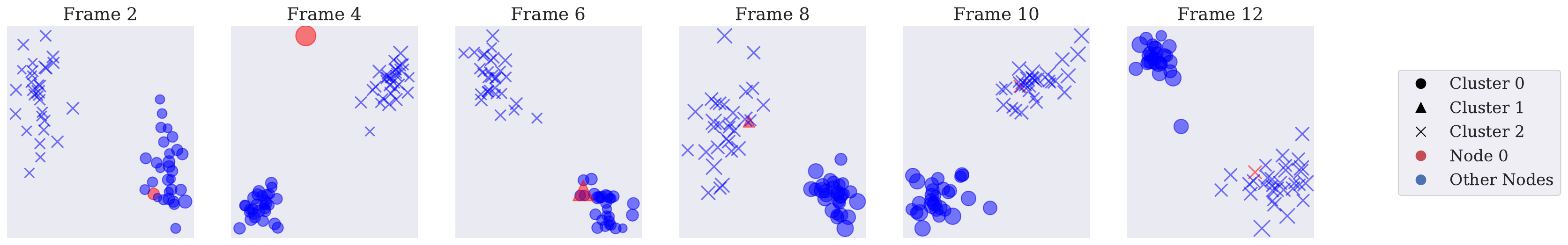}
        \caption{$K=15, \tau=50.0$ 
        }
        \label{fig:sbm:b}
    \end{subfigure}
    \begin{subfigure}{\textwidth}
        \centering
    \includegraphics[width=\linewidth]{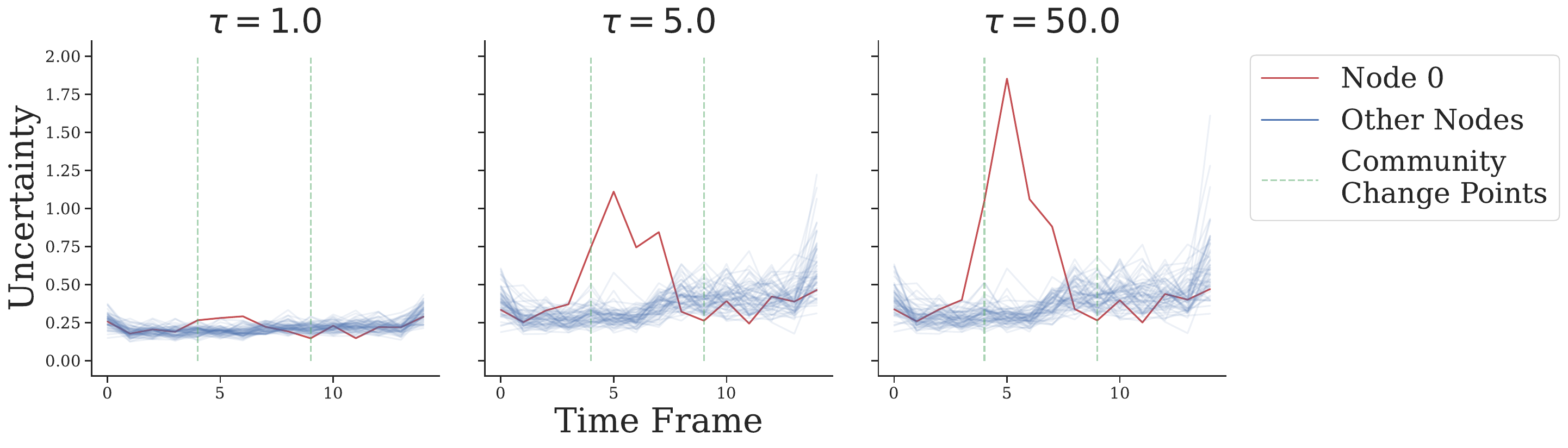}
    \caption{Uncertainty over time for $K=15$ and different values of the prior scale $\tau$.
    The estimated uncertainty increases with the scale parameter.
    Moreover, for a high value of $\tau$, the node changing community will yield a higher variance around its associated embedding position}
    \label{sbm:uncertainty_vs_t}
    \end{subfigure}    
    \caption{Resulting latent positions on the Stochastic Block Model. Uncertainty is represented by the size of the markers.
    In the first period, the nodes are divided into two communities (Circles and crosses).
    Then in the second one, node 0 becomes a triangle and forms its own community. During that transition, node 0's uncertainty increases, especially when using a less informative prior ($\tau=50.0$). Finally, the same node 0 becomes a cross. }
\end{figure*}
\label{experiments}
We performed various experiments to answer the following research questions.
First, we evaluate the uncertainty of the latent positions on simulated data, and on real-world datasets.
Next, we try to understand qualitatively how the parameters of the model affect the resulting latent positions. Finally, we try to understand to what extent the TGNE method allows reconstructing the events of unobserved edges, based on the event history of the observed edges.

\ptitle{Datasets}
In our experiments, we used a simulated dataset, as well as four real-world datasets, for which we provide a brief description below.
The \textbf{HighSchool} dataset\cite{HighSchoolStudents} is a contact network of student in a French preparatory class High School in Marseille. Their interactions were recorded using wearable devices over 9 days.
The resulting embeddings are shown in Figure \ref{Fig:Highschool}.
The \textbf{MIT Reality Mining} Dataset \cite{eagleRealityMiningSensing2006} is a dataset of face-to-face contacts between participants of an experiment ran by members of MIT media Lab. The data was collected over the course of around 9 months, between 2004 and 2005.
The obtained embeddings for this dataset are shown in Figure \ref{Fig:MIT}.
The \textbf{Workplace} dataset is a dataset of face-to-face contacts between employees in a workplace \cite{NWS:9950811}. 
Their interactions were recorded on 11 days (2013/06/24 to the 2013/07/05). 
In this work we focus on the first day of interactions.
The \textbf{UCI} dataset is a Facebook-like, unattributed online communication network among students of the University of California at Irvine, along with timestamps with the temporal granularity of seconds. We used the preprocessed version from the recent DGB Benchmark \cite{dgb_neurips_D&B_2022}.
A summary of the datasets is shown on Table~\ref{table:dataset_stats}, along with the associated runtimes of the TGNE method. 


\begin{table}[ht]
    \centering
    \begin{tabular}{lllll}
        \toprule
        Dataset                   &  HighSchool &     RealityMining &  Workplace &    UCI \\
        \midrule
        Nodes       &         180 &     106 &         92 &   1899 \\
        Unique edges &         758 &    5756 &        755 &  20295 \\
        Events       &        9957 &  779868 &       9827 &  59834 \\
        \midrule
        Runtime (s)               &          45 &     109 &         40 &    542 \\
        \bottomrule
        \end{tabular}
    \caption{Statistics on the Datasets, and associated runtime of TGNE for 500 epochs. 
    }    
    \label{table:dataset_stats}
\end{table}

\subsection{Example on data simulated using a Stochastic Block Model}
\label{sec:exploratory}
We evaluate the estimated uncertainty of the interactions in an example simulated using a Stochastic Block model(SBM), where one node changes community over time, while all the other nodes stay in the same community. 
This data generation procedure is adapted from \cite{rastelliContinuousLatentPosition2021}, but here we focus on the uncertainty aspect.
A detailed explanation of the simulation procedure is provided in the Appendix. 
The resulting Gaussian Embeddings are shown in Figure \ref{fig:sbm:a} and \ref{fig:sbm:b}, for two sets hyperparameters.
Using a low scale parameter, the positions are located with more precision, and the trajectories evolve in a smoother way between time stamps. This is to be expected since the regularization term is stronger in that case. However, the estimated uncertainty is uniform across nodes in that case.
In the high-scale regime, the trajectories evolve more freely between frames, as in that case, the between-frame regularization is weaker. However, the uncertainty of the node that changes community is higher than the uncertainty of the other nodes, as expected. 
On Figure \ref{sbm:uncertainty_vs_t} we show the evolution of the uncertainty of the node that changes community over time, for different value of the scale parameter.

\subsection{Uncertainty evaluation}
\label{sec:uncertainty}
We leverage the probabilistic nature of the TGNE method to analyze the model uncertainty.
\begin{figure}[h]
    \centering
    \includegraphics[width=0.5\linewidth]{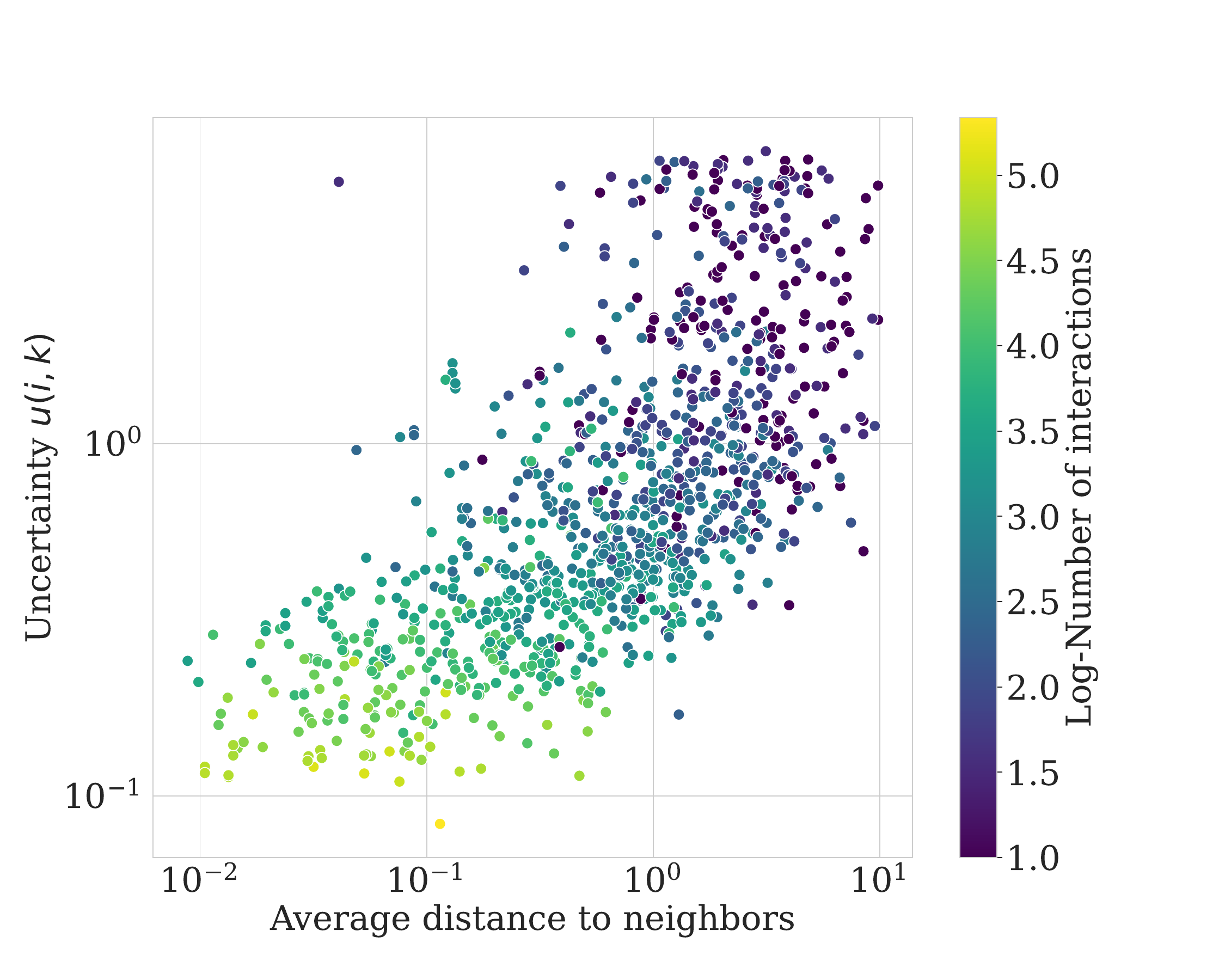}
    \caption{
        Log-log plot of the node-level uncertainty u(i,k) as a function of the average distance to the neighbors whithin the same interval, with ($\tau=50.0, K=15$).
        }
    \label{fig:node_u}
\end{figure}

\ptitle{Node-level uncertainty}
The TGNE method outputs a Gaussian distribution for each node at each individual time stamp. Thus, the uncertainty around the latent positions can be naturally measured through the scale of the variational Normal distribution. While there are multiple potential sources of uncertainty, we empirically assess the impact of the node degree on the uncertainty of the latent positions. 
Namely, for each node $i$ and each sub-interval $I_k$, we measure the uncertainty of node $i$ on interval $I_k$ by calculating $u(i,k) = \frac{\sigma_i^{(k)}+\sigma_i^{(k+1)}}{2}$ and conversely calculate the number of interactions $\mN_{i}(I_k)$ of the node on this interval. 
Moreover, we relate the uncertainty associated with a node on a given interval to the average Euclidean distance to its neighbors on the same interval. In order to display how these different value are related, in Figure \ref{fig:node_u} we represent the average uncertainty $u(i,k)$ as a function of the average distance to the neighbors within the same interval.

\ptitle{Edge-level uncertainty}
The uncertainty about the node's latent positions can be propagated into a notion of uncertainty on the distribution generating the temporal graph, materialized by the posterior predictive distribution defined by the Poisson Processes $\pp(\tilde{\lambda}_{ij})$ with the random variable $\tilde{\lambda}_{ij}$ being defined as:
\begin{align*}
    \tilde{\lambda}_{ij}(t) \deltaequal \E[\lambda_{ij}(t)|\vz]     
\end{align*}

Note that here we get a distribution over the set of joint Poisson Process distributions. 
While evaluating this posterior predictive distribution is intractable, we can approximate it by sampling $B$ i.i.d. samples from the latent code, i.e. $\vz^{(b)}\iidsim q_{\phi}$ for $b=1,\ldots,B$. Then for each sample $\vz^{(b)}$ we denote $\lambda_{ij}^{(b)}$ (respectively $\Lambda_{ij}^{(b)}(I_k)$) the rate function (respectively the cumulative rate) obtained by plugging $\vz^{(b)}$ into the similarity function defined in \ref{sec:clpm}.
In our experiments, we evaluated the predictive uncertainty on the cumulative rate defined as 
$
    Std(\tilde{\Lambda}_{ij}(I_k))\approx \sqrt{\frac{1}{B}\sum_{b=1}^B \parentheses{
        \Lambda_{ij}^{(b)}(I_k) - \E[\tilde{\Lambda}_{ij}(I_k)]
    }^2}
    $
where
$
    \E(\tilde{\Lambda}_{ij}([I_k])) \approx \frac{1}{B}\sum_{b=1}^{B} \Lambda_{ij}^{(b)}(I_k).
$
We calculate, for each unique value of $N_{ij}(I_k)$ the average uncertainty of the cumulative rate over all the edges that have $N_{ij}(I_k)$ interactions in $I_k$.
In our experiments, we found out that the model uncertainty on $\Lambda_{ij}(I_k)$ decreases with the number of interactions for $i,j$ in $I_k$, that we denote here $\mN_{ij}(I_k)$.

\begin{figure}[h!]
	\begin{subfigure}{0.49\linewidth}
		\centering
		\includegraphics[width=\textwidth]{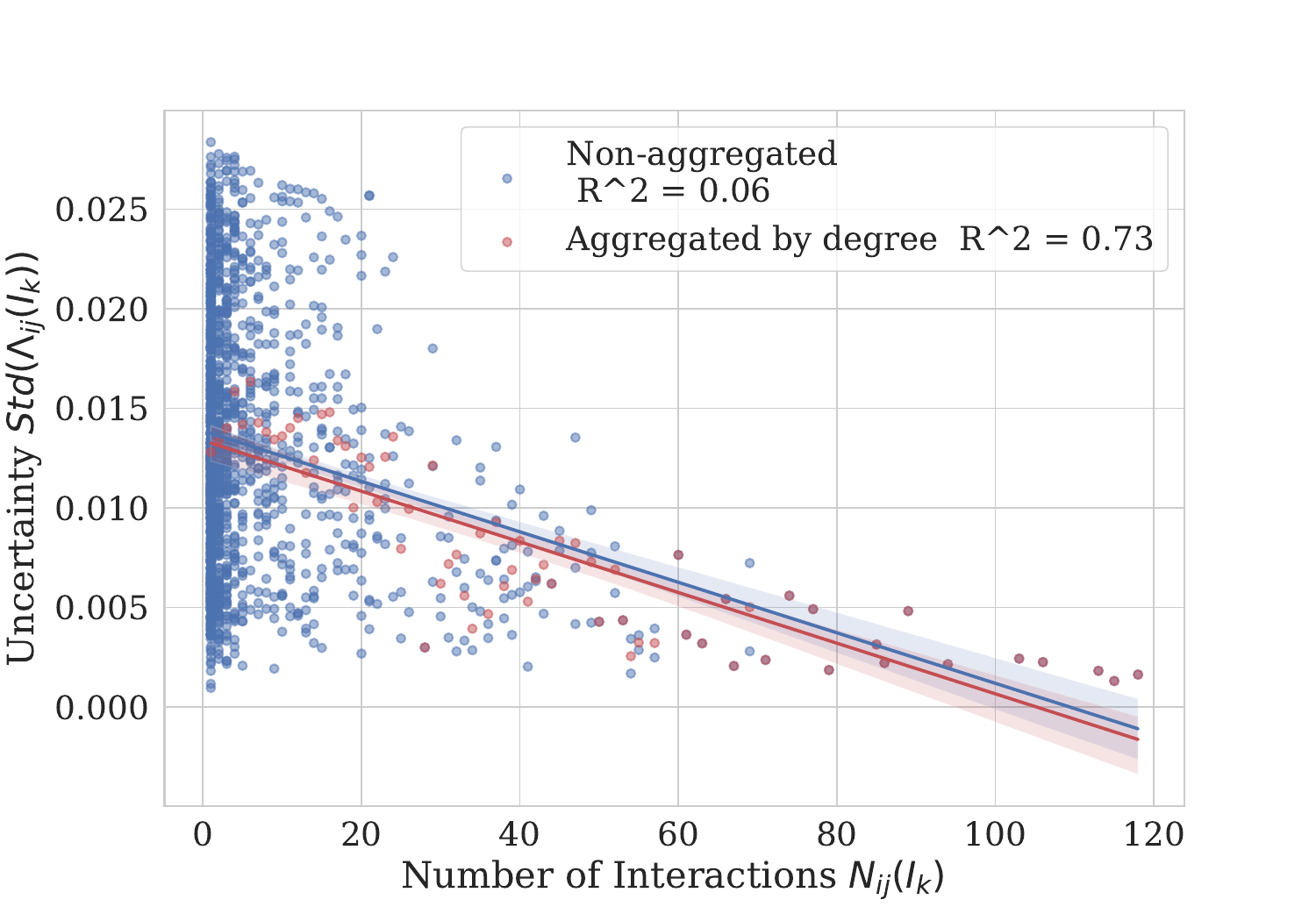}
		\caption{Edge-level uncertainty $Std(\tilde{\Lambda}_{ij}(I_k))$ as a function of $N_{ij}(I_k)$, with ($\tau=1.0, K=15$). }
		\label{fig:edge_u}
	\end{subfigure}	
\hfill
	\begin{subfigure}{0.49\linewidth}
		\centering
		\includegraphics[width=\linewidth]{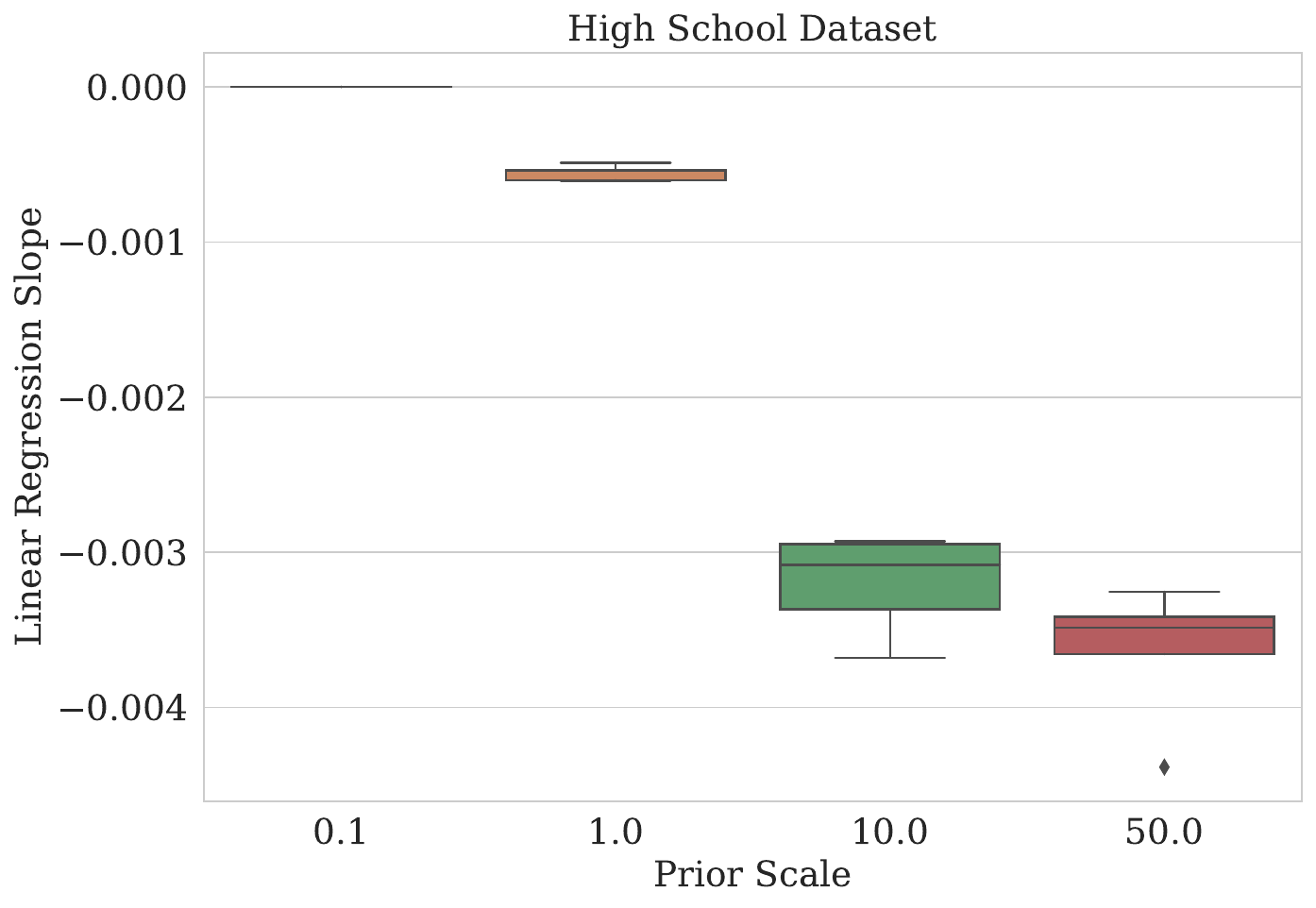}
		\caption{Linear Regression Slope of $Std(\tilde{\Lambda}_{ij}(I_k))$ against $\mN_{ij}(I_k)$ for different values of the prior scale}
		\label{fig:uncertaintylinreg}
	\end{subfigure}	
\hfill
\begin{center}
	\begin{subfigure}{\linewidth}
		\centering
		\includegraphics[width=0.5\linewidth]{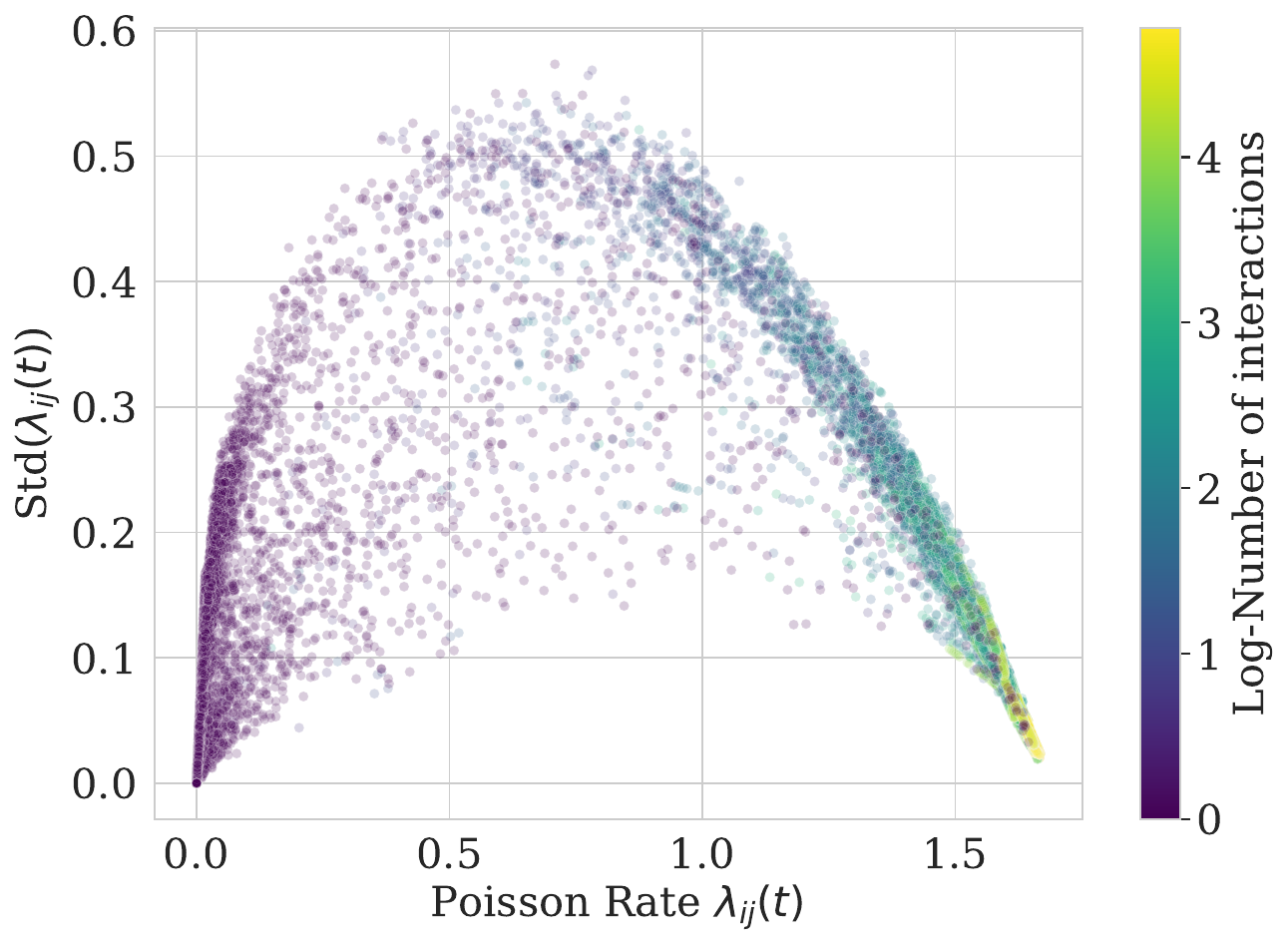}
		\caption{Uncertainty vs Poisson Rate on the High School Dataset with ($K=15, \tau=1.0$). The Poisson Rate is calculated for each positive event and associated negative event. For each event $(u,v,t)$ we color it by the number of interactions between $(u,v)$ in the interval $I_k$ such that $t \in I_k$.
		Events with extreme Poisson Rate values are associated with a low uncertainty, while intermediate Poisson Rates are associated with a higher uncertainty.
		\label{fig:u_vs_score}    
		}
	\end{subfigure}	
\end{center}
    \caption{Edge Level Uncertainty}
\end{figure}

\begin{figure*}
   
    \begin{subfigure}[!]{\textwidth}
        \centering
        \includegraphics[width=1\linewidth]{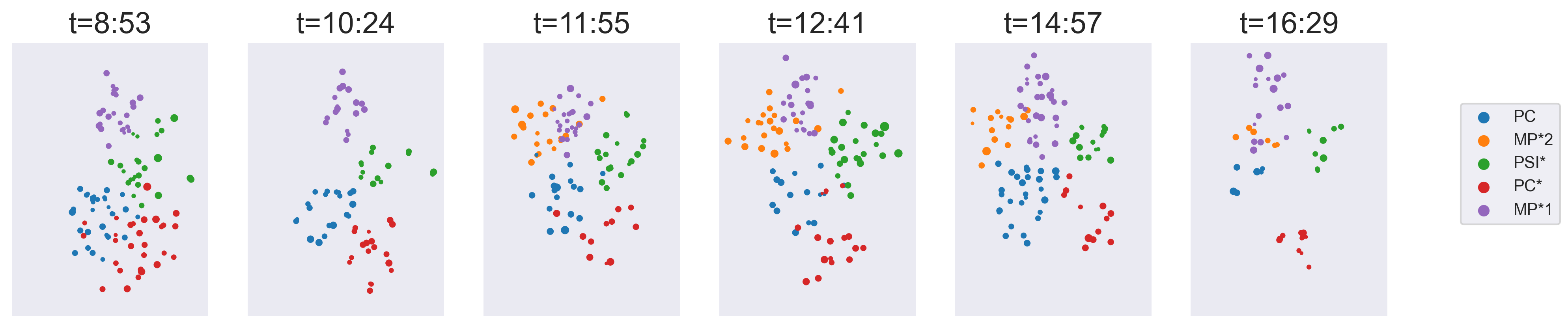}
        \caption{Resulting latent positions on the High School Dataset, with $\tau=1.0, K=15$. Only the nodes that appear during each time frame are shown. The color corresponds to the class of the student, while the size is proportional to the uncertainty on their latent positions. 
        Students from similar classes (e.g. Physics/Chemistry students (PC and PC*) cluster together more than with other students (e.g. Mathematics/Physics: MP*1 and MP* 2)).
        Some students from PSI* (Physics and Engineering Science) still seem to interact with the PC/PC*.}
        \label{Fig:Highschool}
    \end{subfigure}
    \begin{subfigure}[!]{\textwidth}
        \centering
        \includegraphics[width=1.0\linewidth]{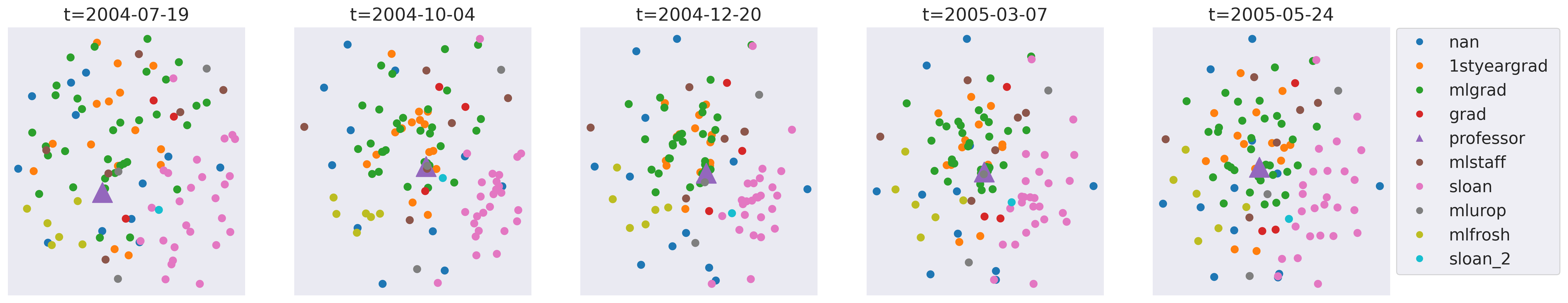}
        \caption{Resulting latent positions on the MIT Reality Mining Dataset, with $K=15, \sigma=0.1$. 
        The embedding positions allow distinguishing members of different departments (the people of the Sloan Business school form a clear separate community). Moreover, the professor (represented by a purple triangle) is in the center of the embedding space and over time moves around between the communities he mostly interacts with.
        }
        \label{Fig:MIT}
    \end{subfigure}    
    \caption{Latent Positions obtained on the Toy Dataset, the Highschool Dataset and the MIT Reality Mining Dataset.}
\end{figure*}

In Figure \ref{fig:uncertaintylinreg}, we observe that the linear regression slope decreases with the prior scale, suggesting that a less informative prior leads to a stronger correlation between uncertainty and the number of interactions.
There is no generic best choice of the regularization parameter, it will depend on the task. It may for example be trained using cross-validation for predictive tasks, while for unsupervised tasks it may be less straightforward to choose it well. 
Its effect is nonetheless evident: it introduces a bias-variance trade-off between concentrating the trajectories in the latent space over time (which would increase the \textbf{bias}) and modeling the observed interactions in time more closely (thus increasing the \textbf{variance}).

\ptitle{Relationship between the uncertainty and the Poisson rate}
In order to visualize the relationship between the Poisson Rate and the learned notion of uncertainty, we select negative samples for each positive event, by swapping the destination node with a random node in the network. 
Then we calculate the Poisson Rate for each positive event and associated negative event, and compare it with the uncertainty propagated from the latent positions to the rate function.
The results are shown in Figure \ref{fig:u_vs_score}. In general, more extreme Poisson Rates seem to be associated with less uncertainty.

\clearpage

\subsection{Temporal Network Reconstruction}
\label{ssec:tnr}
In this experiment we assess in which case the TGNE method is good at reconstructing interactions of unobserved edges.

\ptitle{Task}
The usual setting for evaluating Temporal Graph Embedding method involves splitting the dataset into a train and a test set, using a cutoff point in time, such that the interactions before the cutoff are the train data and after the cutoff the test data. This does not apply here, as the TGNE model is not inductive on the time axis.
However, TGNE is well-suited for \emph{Temporal Network Reconstruction}: reconstructing missing interactions from the continuous-time temporal graph. Here we empirically evaluate it on this task.
We split the edges in the network into train,validation and test sets.
Then for each split, and each interval $I_k$, we predict whether each edge $e=(i,j)$ interacts in the interval $I_k$, i.e. whether there exists an \emph{interaction} $(i,j,t)$ in history, such that $t\in I_k$. For each interval and each positive edge we sample a single negative edge, thus casting the problem into a binary classification task of the node-pair/ interval triplets $(i,j,k)$.

We compare four different approaches for scoring the triplets $(i,j,k)$. For our method, a score is calculated based on a fitted \textbf{TGNE} model, as the expected amount of interaction on the interval: $score(i,j,k) = \Lambda_{ij}(I_k)$.
A first baseline is derived by postulating a binary, Euclidean Latent Space Distance Model (\textbf{LSDM}) \cite{Hoff2002} on the interactions occurring on each interval: $score(i,j,k)  =\sigma(\beta - ||z_i^{(k)}-z_j^{(k)}||^2)$,
where the latent positions $z_j^{(k)}$ are optimized using Maximum Likelihood Estimation.
A second baseline is popularity-based prediction, also named Preferential Attachment (\textbf{PA}): $score(i,j,k) = deg(i,k)\cdot deg(j,k)$ where $deg(i,k)=\sum_j N_{ij}(I_k)$ is the degree of node $i$ on interval $k$.
Finally, we include a Random Baseline (\textbf{Random}), that calculates a random score for each triplet.
In order to discuss more precisely the regularizing effect of the prior for Network reconstruction, techniques such as tensor decomposition could have been explored, however, since TGNE is derived from Latent Space Models, we decided to stick with this class of models in this work.

\ptitle{Results}
For the \textbf{HighSchool} dataset and the \textbf{UCI} dataset we use 10 \% of the edges as test edges, and the rest as train edges. For the \textbf{Workplace} dataset we use 30 percent of the edges as test edges, and the rest as train edges.
The results are provided in Table $\ref{auc_results}$.
On the High School Dataset, it can be seen that while a binary Latent Space Distance Model could be used to predict the presence/absence of links, its resulting configuration of node embeddings overfits the training data, and thus does not perform well on the test data.
In contrast, the embeddings obtained using TGNE perform worse on the training set, but much better on the test set. This showcases the benefits of the regularization term on the predictive abilities of the model.
\begin{table}[ht]
    \centering
    \begin{tabular}{lrrrr}
        \toprule
        {} &   TGNE &      LSDM &        PA &    Random \\
        Dataset    &           &           &           &           \\
        \midrule
        HighSchool(train) &  0.959 &  \textbf{0.999} &  0.904 &  0.497 \\
        HighSchool(test)  &  \textbf{0.885} &  0.784 &  0.756 &  0.524 \\
        Workplace(train) &    0.901 & \textbf{1.000} & 0.882 &   0.497 \\
        Workplace(test)  &    \textbf{0.702} & 0.665 & 0.672 &   0.496 \\
        UCI (train) &    0.972 & \textbf{0.999} & 0.967 &   0.506 \\
        UCI (test)  &    \textbf{0.922} & 0.882 & 0.914 &   0.505 \\ 
        \bottomrule
    \end{tabular}
    \caption{AUC Results on the different datasets for $K=15$ intervals.}
    \label{auc_results}
\end{table}

\section{Discussion and Future Work}
\label{discussion}
In this paper, we discuss the performances of TGNE and its ability to capture the uncertainty on the estimated latent positions and the ability of the obtained locations to predict the occurrence of edges in successive intervals.
However, some open questions remain, which we detail here for future work.

\ptitle{Adapt the changepoints to the density of interactions}
The number $K$ and the positions of the changepoints $\eta_0,\ldots,\eta_K$ are fixed in the TGNE. 
As all the interaction times are re-scaled to be between 0 and 1, the constant step size is fixed to $\eta_{k+1} - \eta_k=\frac{1}{K}$. 
However, adapting the step size to the observed rate of events would naturally produce a more fine-grained representation of the temporal network structure in sub-intervals where more events happen. This appears as a promising avenue to improve model efficiency.

\ptitle{Identifiability}
In the high-scale regime (see for instance Figure \ref{fig:sbm:b}), there is significant rotation of latent configurations from one frame to the next. This is because the model fails to identify the rotations of the configurations. Although this issue is partly mitigated by the effect of the prior, it could potentially be resolved through the use of a Procrustes transform applied to the configurations of trajectories.

\ptitle{Node Inductivity}
The proposed model is transductive, meaning it is limited to the set of nodes that are provided in advance and cannot embed unobserved nodes. In contrast, an alternative approach would be to use amortization, as in seminal works such as \cite{Kipf2016}, to map nodes and their context to Gaussian parameters using a parametric function. This approach would allow predicting trajectories for unobserved nodes, and allow the resulting model to scale to millions of nodes.

\ptitle{Time Inductivity}
As mentioned earlier, the TGNE model could be used to learn dynamics (or distributions thereof) in the embedding space instead of directly learning a sequence of latent distributions in the latent space. This would enable extrapolating the dynamic to future unobserved links. One-step ahead Link Prediction would be a key metric to evaluate the success of such an approach.

\ptitle{Continuous Time Encoder}
Finally, related to the previous point, one limitation of the proposed approach is that it relies on a \emph{discrete time encoder} since each node is essentially mapped to a sequence of Gaussian parameters. However, one alternative approach would be to build on \cite{chenNeuralOrdinaryDifferential2019} to embed the nodes into parameters of a joint stochastic process on the node state and the network state, and using a Point Process Model as a decoder.

\section{Conclusion}


In the present work, we have presented a principled approach to Temporal Graph embedding that leverages Variational inference to infer latent distributions on node trajectories from an observed temporal network. This is in contrast with traditional temporal graph embedding methods, where only a trajectory of points per node is usually estimated. Our results show that in the case where the prior distribution is not restrictive enough, the uncertainty coming from this greater degree of freedom in the latent space can be partially captured back in the scale parameter of the estimated normal distributions. On top of that, the reconstruction experiment showcases the need for regularization in the case of temporal graph embeddings, as it makes the obtained trajectories more easily readable visually, but also leads to better reconstruction results. Finally, estimating uncertain node positions through time could lead to other applications such as anomaly detection.




\appendix






\bibliographystyle{plainnat}
\bibliography{references}  






\section{Supplementary Material}

\subsection{Code}
An implementation of TGNE is provided in the supplementary material.

The datasets can be downloaded from the following urls:
\begin{itemize}
    \item The Reality Mining Dataset can be downloaded \href{http://realitycommons.media.mit.edu/realitymining.html}{here}. The user needs to be authorized before being allowed access to the data.
    \item The High School contact network dataset is publicly available \href{http://www.sociopatterns.org/datasets/high-school-contact-and-friendship-networks/}{here}.
    \item The Workplace dataset can be found \href{http://www.sociopatterns.org/datasets/contacts-in-a-workplace/}{here}.
    \item The UCI dataset can be downloaded from \href{https://zenodo.org/record/7213796#.ZFkv_o3P1cB}{here}.
\end{itemize}

\subsection{Simulated Data}
\label{appendix:toy}
\paragraph*{Generation Procedure}
In this example, a network of 60 nodes (indexed from 0 to 59) is observed during 3 segments of time of equal duration, denoted $I_1,I_2,I_3$. All the nodes are assigned static cluster assignments, except the first one, which goes into its own community on the second segment and then joins the opposite community.
In the first segment, the nodes are split into two clusters: the nodes from 0 to 29 go into cluster $C_0$, while nodes 30 to 59 go into $C_1$.
In the second segment, node 0 goes into its own cluster, say $C_2$, while the other nodes stay in their respective clusters. In the third segment, node 0 goes into the cluster $C_1$, thus only two clusters are present at this time.
Based on these cluster assignments, we generated the interactions using a Stochastic Block Model: we fix inter and intra-cluster interaction rate, and proceed as follows.
For each segment$s$ and each node pair $(i,j)$, let $r(i,j,s)$ be the rate of interaction between $i$ and $j$ during segment $s$, as defined by the SBM.
We first generate a number of interactions $N(i,j,s)\sim Poisson(r(i,j,s))$, and then sample $N(i,j,s)$ timestamps uniformly distributed over the time segment $I_s$.
\paragraph*{Results}
On the simulated data, we start by fitting a model with 15 changepoints and scale parameters $\sigma=1.0, \sigma_0=1.0$.
The nodes are colored by cluster label and the size is proportional to the log of their number of interaction in a small time window aound each changepoint.
In the first frame of Figure \ref{fig:sbm}, the nodes are grouped into two clusters. Then on frame 10, the first node (in orange), leaves the blue community and goes into its own cluster. Finally, the orange node joins the green community at frame 10.

The uncertainty over time for different values of the hyperparameters is shown on Figure 4.

\subsection{Calculation of the Cumulative Rate on an interval where the trajectories are linear}
\label{proof:cumulative_rate}
Let's calculate the integral $\int_{\eta_{k-1}}^{\eta_k}\lambda_{ij}(s)ds.$
Under the piece-wise linear assumption, the trajectory of node $i$ at time $s=((1-t)\eta_{k-1}+t\eta_k)\in I_{k}$ (with $t\in [0,1]$) can be written as:

$z_i((1-t)\eta_{k-1}+t\eta_{k})= (1-t)z_i(\eta_{k-1})+tz_i(\eta_{k})$.
Based on that, let's rewrite the rate function in a way that makes it easier to integrate.

The log of the rate writes:
\begin{align*}
    log \lambda_{ij}(s) 
    &= \beta - ||z_i(s) - z_j(s)||^2  \\
    &= \beta - \gamma_{ij}(s)  \\
\end{align*}

where, denoting $\Delta_{ij}(\eta_k) = z_i(\eta_k) - z_j(\eta_{k})$, $\gamma_{ij}$ is defined as 
\begin{align}
    \gamma_{ij}((1-t)\eta_k+t\eta_{k+1}) 
    &= ||(1-t)\Delta_{ij}(\eta_k) + t\Delta_{ij}(\eta_{k+1})||^2.    
    \label{gamma_ij}
\end{align}

In particular, $\gamma_{ij}$ is a second-order polynomial in $t$. 
Our goal now is to express $\gamma_{ij}$ as the log of the density of a normal distribution.

More precisely, let's try to write it under the form

\begin{equation}
    \gamma_{ij}(s) = a + \frac{(t-\mu)^2}{2\sigma^2}  
    \label{eq:canonical}
\end{equation}
for some coefficients $a, \mu, \sigma$, where $s = (1-t)\eta_{k-1}+t\eta_k$

On the one hand, developing the expression \ref{gamma_ij} yields:
\begin{align*}
    \gamma_{ij}(s) = 
    &t^2
    \left[
        ||\dprev||^2 + ||\dnext||^2 - 2\langle\dprev,\dnext\rangle
    \right]\\
    + &t 
    \left[
        -2||\dprev||^2+ 2\langle\dprev, \dnext\rangle
    \right]\\
    + &
        ||\dprev||^2\\
\end{align*}

On the other hand, developing equation \ref{eq:canonical} yields:

$$\gamma_{ij}(s) = a + t^2 (\frac{1}{2\sigma^2}) + t (-\frac{\mu}{\sigma^2}) + \frac{\mu^2}{2\sigma^2}$$

Identifying the coefficients of the polynomial, we get the following system of equations:

\begin{align}
    \frac{1}{2\sigma^2}
    &=\norm{\dprev-\dnext}^2\\
    \frac{\mu}{2\sigma^2}
    &=\langle\dprev, \dprev-\dnext\rangle\\
    a+\frac{\mu^2}{2\sigma^2}&=||\dprev||^2
\end{align}

Finally, solving the system for $a$, $\mu$ and $\sigma$ yields:

\begin{align*}
    \sigma&=\frac{1}{\sqrt{2}\norm{\dprev-\dnext}}\\
    \mu&=\frac{\dotproduct{\dprev}{\dprev-\dnext}}{\norm{\dprev-\dnext}^2}\\
    a&=||\dprev||^2-
    \frac{\dotproduct{\dprev}{\dprev-\dnext}^2}{
        \norm{\dprev-\dnext}^2
    }
\end{align*}

We can conclude by using two changes of variables:
\begin{align*}
    &\int_{\eta_{k-1}}^{\eta_k}\lambda_{ij}(s)ds\\
    &=\int_{\eta_{k-1}}^{\eta_k}\exp(\beta-\gamma_{ij}(s))ds 
    \\
    &=\exp(\beta)(\eta_k-\eta_{k-1})\int_{0}^{1}\exp(-
    \parentheses{
        a+\frac{(t-\mu)^2}{2\sigma^2}
    }
    )dt \tag*{($s = (1-t)\eta_{k-1}+t\eta_k$ such that $ds = (\eta_k-\eta_{k-1})dt$)}
    \\
    &=\exp(\beta-a)(\eta_k-\eta_{k-1})\sigma
    \int_{-\frac{\mu}{\sigma}}^{\frac{1-\mu}{\sigma}}\exp(-
        \frac{u^2}{2}
    )du \tag*{Where we set $u=\frac{t-\mu}{\sigma}$ such that $du=\frac{dt}{\sigma}$}\\ 
    &=\exp(\beta - a)(\eta_{k-1}-\eta_{k}) \sigma \sqrt{2\pi} [\Phi(\frac{1-\mu}{\sigma}) - \Phi(-\frac{\mu}{\sigma})].
\end{align*}

\subsection{KL divergence between a Gaussian Markov Chain and a product of independant Gaussians}
\label{klproof}
Let $x_1,...,x_T$ be some random variables and the two distributions $q$ and $p$ defined as:

\begin{align*}
    q(x) = \prod_{t=1}^T q_t(x_t)
\end{align*}
\begin{align*}
    p(x) = p(x_1)\prod_{t=2}^T p_t(x_t|x_{t-1})
\end{align*}
Then 
\begin{align}
    \label{kl_qp}
    KL(q||p) = KL(q_1, p_1) + \sum_{t=1}^{T-1} \E_{x_t \sim q_{t}}[KL(q_{t+1}||p_{t+1}(.|x_t))]    
\end{align}

In particular, when $$q_t(x_t)=\Ncal(x_t;\mu_t, \sigma_t^2\mI_d)$$ and 
$$p(x) = \Ncal(x_1;\nu_1, \tau_1^2\mI_d)\prod_{t=2}^T \Ncal(x_t;x_{t-1}, \tau_t^2\mI_d)$$

$$p_1(x_1) = \Ncal(x_1;\nu_1, \tau_1^2\mI_d)$$
and $$p_t(x_t|x_{t-1})= \Ncal(x_t;x_{t-1}, \tau_t^2\mI_d)$$
Moreover, the KL Divergence between two $d$-dimensional Gaussian distributions is given by:

\begin{align*}
    KL(\Ncal(\mu_1, \sigma_1^2\mI_d)||\Ncal(\mu_2, \sigma_2^2\mI_d))
    &= 
    \frac{||\mu_2-\mu_1||^2}{2\sigma_2^2} \\
    &+ d\left[\log(\frac{\sigma_2}{\sigma_1}) + \frac{\sigma_2^2}{\sigma_1^2}-\frac{1}{2}\right]
\end{align*}

this yields:
\begin{align*}
    KL(q_1|| p_1) = \frac{||\mu_1-\nu_1||^2}{2\tau_1^2} 
    + d\left[\log(\frac{\tau_1}{\sigma_1}) + \frac{\tau_1^2}{\sigma_1^2}-\frac{1}{2}\right]
\end{align*}
\begin{align*}
    KL(q_t|| p_t(.|x_{t-1})) = \frac{||x_{t-1}-\mu_t||^2}{2\tau_t^2} 
    + d\left[\log(\frac{\tau_t}{\sigma_t}) + \frac{\tau_t^2}{\sigma_t^2}-\frac{1}{2}\right]
\end{align*}
so
\begin{align*}
    &\E_{x_{t-1}\sim q_{t-1}}\left[
        KL(q_t|| p_t(.|x_{t-1}))
    \right] \\ 
    &=  d\left[\log(\frac{\tau_t}{\sigma_t}) + \frac{\tau_t^2}{\sigma_t^2}-\frac{1}{2}\right]\\
    &+\E_{x_{t-1}\sim q_{t-1}}\left[ \frac{||x_{t-1}-\mu_t||^2}{2\tau_t^2} \right]\\
    &=d\left[\log(\frac{\tau_t}{\sigma_t}) + \frac{\tau_t^2}{\sigma_t^2}-\frac{1}{2}\right]\\
    &+\frac{1}{{2\tau_t^2}}\E_{x_{t-1}\sim q_{t-1}}\left[ ||x_{t-1}-\mu_{t-1}||^2+||\mu_{t}-\mu_{t-1}||^2 \right]\\
    &=d\left[\log(\frac{\tau_t}{\sigma_t}) + \frac{\tau_t^2}{\sigma_t^2}-\frac{1}{2}\right]+\frac{1}{{2\tau_t^2}}\left[||\mu_{t}-\mu_{t-1}||^2+ \sigma_{t-1}^2 \right]\\
\end{align*}

So finally we get 

\begin{align*}
    KL(q||p) 
    &= 
    \frac{||\mu_1-\nu_1||^2}{2\tau_1^2} \\
    &+k\sum_{t=1}^T
    \left[
        \log(\frac{\tau_t}{\sigma_t}) 
        + 
        \frac{\tau_t^2}{\sigma_t^2}-\frac{1}{2}
        \right]\\
    &+\sum_{t=1}^T 
    \frac{
        ||\mu_{t}-\mu_{t-1}||^2+ \sigma_{t-1}^2
    }{2\tau_t^2}\\
\end{align*}


\subsection*{Runtime comparison}
\begin{figure}[h!]
	\centering
    \includegraphics[width=0.5\linewidth]{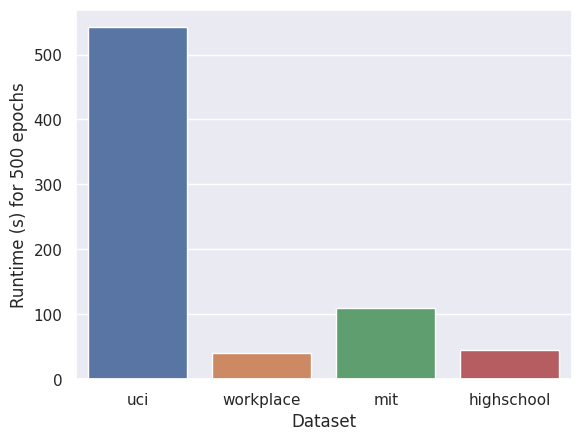}
   \caption{Runtime of TGNE for 500 epochs. 
   We ran all the experiments on a server with a 12 Core Intel(R) Xeon(R) Gold processor and 256 GB of RAM. 
   The method scales well with the number of interactions. However, the number of unique edges has a bigger impact on the runtime.}
\end{figure}

\end{document}